\documentclass{article}

    \PassOptionsToPackage{numbers, compress}{natbib}
\usepackage[preprint]{neurips_2023}

\usepackage{amsmath} 
\usepackage[utf8]{inputenc} % allow utf-8 input
\usepackage[T1]{fontenc}    % use 8-bit T1 fonts
\usepackage{hyperref}       % hyperlinks
\usepackage{url}            % simple URL typesetting
\usepackage{booktabs}       % professional-quality tables
\usepackage{amsfonts}       % blackboard math symbols
\usepackage{nicefrac}       % compact symbols for 1/2, etc.
\usepackage{microtype}      % microtypography
\usepackage[dvipsnames]{xcolor}         % colors
\usepackage{graphicx}
\usepackage{multirow}

\title{AlphaBlock: Embodied Finetuning for \\ Vision-Language Reasoning in Robot Manipulation}

\newcommand*{\affaddr}[1]{#1} % No op here. Customize it for different styles.
\newcommand*{\affmark}[1][*]{\textsuperscript{#1}}
\newcommand*\samethanks[1][\value{footnote}]{\footnotemark[#1]}

\author{%
Chuhao Jin\affmark[1]\thanks{Equal Contribution. This work was performed when Chuhao Jin, Wenhui Tan and Jiange Yang were visiting Microsoft Research as research interns.}, Wenhui Tan\affmark[1]\samethanks[1], Jiange Yang\affmark[2]\samethanks[1], Bei Liu\affmark[3]\thanks{Corresponding authors: Bei Liu (bei.liu@microsoft.com), Jianlong Fu (jianf@microsoft.com).}, Ruihua Song\affmark[1],
Limin Wang\affmark[2], Jianlong Fu\affmark[3]\samethanks[2]\\
\normalsize
\affaddr{\affmark[1]Renmin University of China},
\affaddr{\affmark[2]Nanjing University},\\
\affaddr{\affmark[3]Microsoft Research}
}

\begin{document}

\maketitle

\begin{abstract}
We propose a novel framework for learning high-level cognitive capabilities in robot manipulation tasks, such as making a smiley face using building blocks. These tasks often involve complex multi-step reasoning, presenting significant challenges due to the limited paired data connecting human instructions (e.g., making a smiley face) and robot actions (e.g., end-effector movement). Existing approaches relieve this challenge by adopting an \textbf{open-loop} paradigm decomposing high-level instructions into simple sub-task plans, and executing them step-by-step using low-level control models. However, these approaches are short of instant observations in multi-step reasoning, leading to sub-optimal results. To address this issue, we propose to automatically collect a cognitive robot dataset by Large Language Models (LLMs). The resulting dataset \textbf{AlphaBlock} consists of $35$ comprehensive high-level tasks of multi-step text plans and paired observation sequences. To enable efficient data acquisition, we employ elaborated multi-round prompt designs that effectively reduce the burden of extensive human involvement. We further propose a \textbf{closed-loop} multi-modal embodied planning model that autoregressively generates plans by taking image observations as input. To facilitate effective learning, we leverage MiniGPT-4 with a frozen visual encoder and LLM, and finetune additional vision adapter and Q-former to enable fine-grained spatial perception for manipulation tasks.    
We conduct experiments to verify the superiority over existing open and closed-loop methods, and achieve a significant increase in success rate by 21.4\% and 14.5\% over ChatGPT and GPT-4 based robot tasks. 
Real-world demos are shown in \small{\url{https://www.youtube.com/watch?v=ayAzID1_qQk}}.
\end{abstract}

\section{Introduction}

Learning high-level cognitive capabilities in robot manipulation tasks is a critical area of research in robotics \cite{trends,cognition}. These tasks pose grand challenges that require robots to understand and execute complex language instructions involving perception, reasoning and manipulation. For example, consider the task of ``making a smiley face with building blocks''. To accomplish this goal, robots need to perceive and identify various building blocks, understand and infer the spatial arrangement necessary to form a smiley face, and precisely manipulate blocks to achieve the desired outcome. This task demands a combination of visual perception \cite{vit,he2016deep}, spatial reasoning \cite{venkatesh2021spatial}, and fine-grained motor control \cite{hiveformer}. Moreover, the challenge lies not only in executing the physical actions but also in comprehending and interpreting the high-level language instructions provided by humans. The success of designing such cognitive robots holds tremendous potential for a wide range of applications in household robots \cite{tidybot}, manufacturing \cite{towards}, and healthcare \cite{health}.  

\begin{figure}[h]
    \centering
    \includegraphics[width=\linewidth]{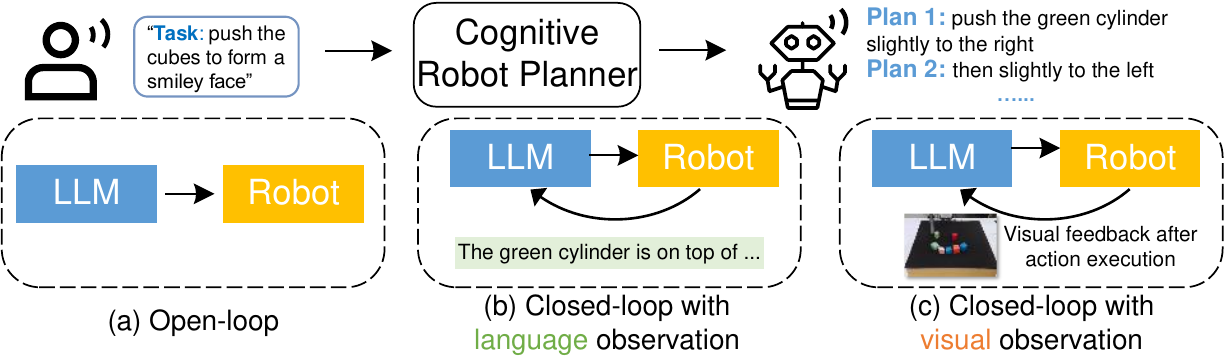}
    \caption{Planner model paradigms. (a) Open-loop models (SayCan-style \cite{saycan}) conduct planning and control separately. (b) Closed-loop models update plans with observation in language (Text2Motion-style \cite{text2motion}).  (c) We infuse more fine-grained visual observation into LLM to update planning.}
    \label{fig:model_pipeline}
\end{figure}

To address the above challenges, recent approaches have turned to using state-of-the-art Large Language Models (LLMs) such as ChatGPT \cite{ouyang2022training} or PALM \cite{chowdhery2022palm} in either open or closed loops. In particular, open-loop approaches \cite{Grounded,saycan} (as shown in Figure \ref{fig:model_pipeline}(a)) utilize LLMs as offline planners to decompose high-level instructions, such as ``making smiley face'', into sub-task plans like ``push one block to a specific position''. They then employ off-the-shelf robot control models \cite{RT1,lynch2022interactive} or APIs \cite{chatgpt} for executing the actions. More recent works \cite{text2motion,chatgpt} close the loop to update the plan based on language-based feedback as in Figure \ref{fig:model_pipeline}(b). Compared with language-based observation which uses symbolic description (e.g., ``on [red, table]'' in Text2Motion \cite{text2motion} means a red cube is on the table), visual observation can depict more comprehensive and fine-grained spatial arrangement of different objects, and thus help to generate more accurate control signals for complex manipulation tasks. As neglecting multi-modal interaction between text-based task description and visual-based observation, existing methods miss out on opportunities to improve overall performance and efficiency.

In this paper, we propose an innovative multi-modal robot control framework, to fulfill high-level cognitive tasks. This framework comprises a newly-collected robot cognitive dataset and a multi-modal plan generation model. Specifically, we name the framework \textbf{CogLoop}, as we aim to bridge the cognitive loop between task planning and control in manipulation tasks. To enable \textbf{end-to-end training}, we require high-quality data triplets consisting of 1) high-level tasks, 2) sub-task plans, and 3) action-observation pairs in robot tasks. However,  the collecting process for such data is time-consuming and expensive. To address this problem, we employ a two-step approach. First, we utilize LLMs, specifically GPT-4 \cite{openai2023gpt4} in our paper, with multi-round prompts to automatically generate a sequence of sub-task plans. This step significantly reduces the reliance on human involvement.
Second, we adopt a state-of-the-art execution model \cite{lynch2022interactive} to generate robot-executable actions based on sub-task plans. Such design further minimizes the need for human intervention, as it automates the generation of actions. Only minimal human efforts are required to select the correct final block layouts. The resulting dataset \textbf{AlphaBlock} encompasses $26$ letters and $9$ fundamental layouts such as lines, triangles and smiley faces formed by building blocks. More details can be found in Section \ref{subsec:data_collection}. 

During training, we build upon recent advancements in multimodal learning that integrate text and image modalities \cite{palme,blip}. Our proposed unified approach incorporates high-level text instructions and image observations to generate sub-task plans. To ensure accurate perception for block characters, such as color and spatial position, we design a vision adapter that extracts and merges multi-stage visual features from the ViT \cite{vit} model in MiniGPT-4 \cite{minigpt4}. To ensure consistent embedding in LLMs, we further employ a visual Q-former \cite{blip} with a language-specific projector, to effectively align image observations with LLMs. The sub-task plans generated are then fed into an off-the-shelf execution model \cite{lynch2022interactive} to predict executable actions, such as offsets along $x$ or $y$ coordinates. 
By leveraging the auto-regressive nature of the holistic framework, encompassing both plan and action generation models, we facilitate mutual reinforcement between task planning and robot behavior. 

\textbf{The contributions} of this paper are summarized as follows: 1) we introduce a novel robot dataset that tackles challenging high-level perception and reasoning tasks using language instructions in robotics; 2) we propose a multi-modal robot control model based on advanced multi-modal foundation models, incorporating a vision adapter specifically designed for robotics tasks; 3) we conduct extensive evaluations and  demonstrate our superior performance over existing competitive robot planning methods. In summary, through seamlessly enabling robots to understand and execute high-level instructions, our framework \textbf{CogLoop} holds significant promise for enhancing cognitive capabilities and facilitating more intelligent interaction with the physical world. 
\section{Related Works}

\subsection{Instruction Following Robot Control}

Creating embodied agents that are capable of general instructions has consistently been an active research field in recent years \cite{affordances,matters,calvin}. Noteworthy advancements have been made by various approaches. 
Hiveformer \cite{hiveformer} employs a history-based approach to enhance motor control by tracking the full history of observation-action pairs. Perceive-actor~\cite{perceiver} encodes language goals and RGB-D voxel observations with a Perceiver Transformer to provide a strong structural prior. 
BC-Z \cite{bcz} and RT-1 \cite{RT1} focus on scaling and expanding the collection of real-world data to facilitate generalization of robots. The field has also benefited greatly from significant progress in multi-modal foundation models like CLIP \cite{clip}, which have served as a catalyst for more advancements. InstructRL \cite{instruction} employs a pre-trained multi-modal autoencoder Transformer \cite{m3ae} to encode instructions and observations, while PAFF \cite{policy} utilizes the pre-trained foundation models CLIP \cite{clip} to provide feedback for relabeling demonstrations. In contrast to previous works that primarily focus on evaluating simple manipulation instructions such as ``picking and placing objects'' \cite{RT1,calvin}, we deal with high-level cognitive instructions in this work, such as ``arrange building blocks as a smiley face.'' Consequently, the transformation of these high-level instructions into executable sub-task plans becomes crucial.

\subsection{Robot Task Planning with LLMs}

Robot task planning refers to the process by which 
a robot generates sub-task plans based on high-level task instructions provided by humans. Recently, several works \cite{palme,Grounded,Monologue,saycan,codex,text2motion,chatgpt} have explored to leverage large language models (LLMs) to plan feasible tasks for robots. SayCan \cite{saycan} generates plans by utilizing action affordances but assumes faultless motor skill execution, which makes it less resilient to intermediate failures due to its open-loop approach. Grounded Decoding \cite{Grounded} jointly decodes the token probability of LLM and grounding functions such as affordances, safety, and preferences. Inner Monologue \cite{Monologue} incorporates grounded feedback from the environment into the LLM to update the planning process. ChatGPT for Robotics \cite{chatgpt} establishes a simple high-level function library and actual APIs for ChatGPT \cite{ouyang2022training} on preferred platforms. PaLM-E \cite{palme} equipped with ViT-22B \cite{22b} and PaLM-540B \cite{chowdhery2022palm} utilizes extensive manually collected robotic data for embodied training. However, such a large number of parameters is not conducive to deploy real-time planning on real robots. In contrast, our CogLoop is equipped with the most popular and lightweight LLM LLaMA \cite{llama} currently available in the open-source community, and conduct parameter-efficient embodied tuning with our AlphaBlock dataset.
\def\modelname{CogLoop}

\section{Approach} \label{sec:approach}
In this section, we first provide a comprehensive description of our problem setting in Section \ref{problem}. Subsequently, we elaborate on dataset collection pipelines for unified planning and control in Section \ref{subsec:data_collection}. Finally, we introduce the carefully-designed architecture and training methodology of our proposed robotic sub-task generation model, referred to as \textbf{CogLoop}, in Section \ref{subsec:model_architecture}.

\subsection{Problem Setup}\label{problem}
Our goal is to design a framework for building cognitive agents capable of conducting high-level manipulation tasks.
To achieve this goal, we first collect an offline dataset $\boldsymbol{\mathcal{D}}$.
Each item in the dataset consists of one high-level task instruction $l$, followed by a series of episodes, where each episode represents one low-level sub-task plan $\mathbf{p}$, and a sequence of observation-action pairs $(\mathbf{o},\mathbf{a})$.
Thus, each training sample can be formed as $\left \{ l, p_{k}, (o_{k,j},a_{k,j})\right \} \sim  \boldsymbol{\mathcal{D}}$.
Here the range of episodes reaches $m$, and the range of observation-action pairs in one episode reaches $n$.
With the collected dataset $\boldsymbol{\mathcal{D}}$, we aim to train a model $\pi_{\theta} (p_{k} \mid l, o_{k, j})$, parameterized by $\theta$.

\subsection{Automatic Data Collection for Real-time Planning and Control} \label{subsec:data_collection}
\begin{figure}[h]
    \centering
    \includegraphics[width=\linewidth]{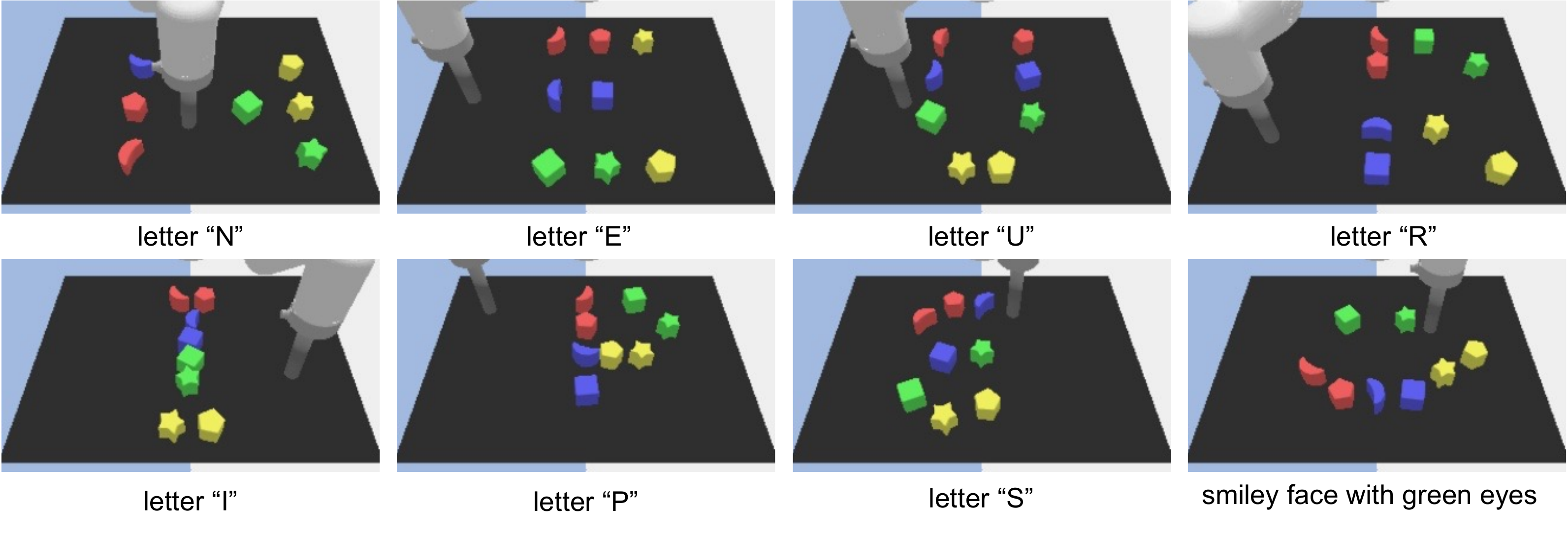}
    \caption{Examples of block placement for high-level task from our AlphaBlock dataset. We show each capital letter of ``NeurIPS'', and ``a smiley face''. The robot arm is placed at random positions.}
    \label{fig:data_demo}
\end{figure}

In order to collect real-time planning and control data for cognitive-level tabletop manipulation tasks, an intuitive approach is to create plans for each scenario and task while maintaining constant interaction with the robots. However, this method is time-consuming and costly, making it challenging to scale up. Consequently, traditional data collection for high-level cognitive robot tasks often focuses on either accumulating plans for diverse tasks\cite{saycan} or gathering low-level robotic actions\cite{lynch2022interactive}, leading to a fragmented approach that may result in sub-optimal outcomes. For instance, a well-devised plan may still lead to task failure if the robot model cannot effectively understand or execute it.

Large language models~(LLM) demonstrate powerful cognitive abilities. An intuitive method is to use LLM for planning and then execute with a robot model. However, due to the accumulation of errors in this method, experimental results show that the success rate of this method is quite low, resulting in low collection efficiency. To tackle this problem, we design a backtracing strategy to effectively gather unified planning and robotic action data in real time by employing LLM (i.e., GPT-4) and a fundamental robotic model, to collect a building blocks manipulation dataset in a tabletop environment, which we have named \textbf{AlphaBlock}. To collect this dataset, we take the following steps:

\paragraph{Task Definition} Initially, we define a set of 35 tasks that primarily involve moving at most eight building blocks to create specific layouts, such as ``a smiley face'' or ``letter Q''. These tasks require both high-level cognitive planning and precise control for moving building blocks. Specifically, the 35 tasks in AlphaBlock can be broadly categorized into four distinct families:
\begin{itemize}
\setlength{\itemsep}{0pt}
\setlength{\parsep}{0pt}
\setlength{\parskip}{0pt}
\item \textbf{Alphabet}: This category covers all 26 uppercase English letters.
\item \textbf{Mathematical Geometries}: Five tasks are designed based on mathematical geometries, including triangles, circles, squares, horizontal lines, and vertical lines.
\item \textbf{Semantic Geometries}: This family includes two geometries that visually represent semantics, such as smiley faces and smiley faces with green eyes.
\item \textbf{Sort by Color or Shape}: The successful completion of these two tasks depends on the model's ability to understand a specific attribute, such as color or shape, and organize the building blocks accordingly.
\end{itemize}

\paragraph{Prompt Design for Placement Collection} We design a prompt for GPT-4 to collect the placement positions of the building blocks to certain layouts, for example, the placement of each block which forms a ``smiley face''. 
Intuitively, we design the prompt as the following format:
\begin{equation*}
<Scene><Task><Rules><Output~Restriction>.
\end{equation*}
We describe the context in $Scene$, such as the size, color, and shape of each building block, orientation of the table, and coordinates of the boundaries. In $Task$, we outline the specific task assigned to GPT-4, i.e., predicting the position of building blocks when assembling a specified layout. 
In $Rules$, we design certain guidelines, such as to avoid collision, to guarantee enough scale of the output layout in the table, etc.
Experimental findings indicate that when merely requesting GPT-4 to provide coordinates, we often get unsatisfactory results. We hypothesize that this deficiency may be attributed to the limited spatial perception inherent within the text-based GPT-4 model. 
To address this issue, inspired by the key insight of chain-of-thought~\cite{COT2022}, in the $Output~Restriction$ section, we require GPT-4 to conduct detailed reasoning before providing specific coordinates, i.e., the reply is limited as the following format: 
\begin{equation*}
<Description><Explain><Positions>.
\end{equation*}
Specifically, we require the GPT-4 model to provide a perceptual understanding and comprehensive description of the given layout within the $Description$ section. For example, it should describe the elements~(lines, curves, and dots) constituting a ``smiley face'', and the positional relationships between these elements. Following this, we request GPT-4 to illustrate, in the $Explain$ section, how to assemble these elements using building blocks. We mandate GPT-4 to explicitly present the relative directions and the Euclidean distances between each building block, thus enhancing its spatial awareness abilities. Ultimately, we ask GPT-4 to produce the coordinates of each building block in the $Positions$ section. To ensure the quality of collected data, we manually check the generated layout and keep about 25\% results. Examples of our AlphaBlock dataset is shown in Figure~\ref{fig:data_demo}. We show our prompts in Appendix~\ref{app:layout}.

To unify the collection of real-time planning and control data, we randomly disrupt at most three positions of these building blocks collected aforementioned, and we aim to move them back to their original positions.
Specifically, we prompt the GPT-4 model to provide a real-time plan and utilize a robotic model LAVA~\cite{lynch2022interactive} to generate the corresponding controlling data.
We design the prompt for the GPT-4 model in the following format: 
\begin{equation*}
<Scene><Task><Plan><Rules><States><Output~Restriction>,
\end{equation*}
where $Scene$ and $Task$ primarily describe the context and the task, and $Rule$ provides guiding principles. In $Plan$ section, we outline the format of the plan, such as which word could be used, and present some plan examples. $States$ encompasses the current positions of the building blocks and the robotic arm, as well as historical plannings.
$Output~Restriction$ indicates the output format for GPT-4 should be as follows:
\begin{equation*}
<Description><Explain><Plan>.
\end{equation*}
We require GPT-4 to describe its understanding of the real-time planning task in $Description$, for example, using numerical values to describe the understanding of current states, boundaries and orientations. In $Explain$, we require GPT-4 to provide a detailed explanation of the reasonableness to formulate the current plan, and outline how to avoid collision with other building blocks. We also require GPT-4 to explicitly provide the relative direction and Euclidean distance given target positions. 
Lastly, a plan should be provided in $Plan$. The LAVA model will execute 12 actions based on this plan, and then we let GPT-4 to generate a new real-time plan based on current observation. We finally determine whether the move task is successful based on the Euclidean distance between the current and target positions. 
To balance the call cost of GPT-4 API and success rate, we only collect data that was successful within 15 plans. We show our prompts in Appendix~\ref{app:coll_real}.

\paragraph{Self-Verification to Ensure Data Quality} Based on a multi-turn conversation principle, we develop a self-verification approach for GPT-4 to automatically check its responses. 
Specifically, regardless of the initial response generated by GPT-4, we pose the question ``Are you sure your answer is correct?'' to prompt the GPT-4 model to check previous answer and provide a new response. Through an examination of various cases, we discover that the overall quality of the second response is generally higher than the first. 
Consequently, we utilize the second response provided by GPT-4 as the final result. We also manually inspect the correctness of the building block placements.

When calling GPT-4 model, we adjust the temperature and ``Top-P'' hyper-parameters to increase the diversity of position and layout. 
At last, we collect 1,345 layouts and 10,669 successful plans for all the 35 tasks.

\begin{figure}[t]
\centering
\includegraphics[width=\textwidth]{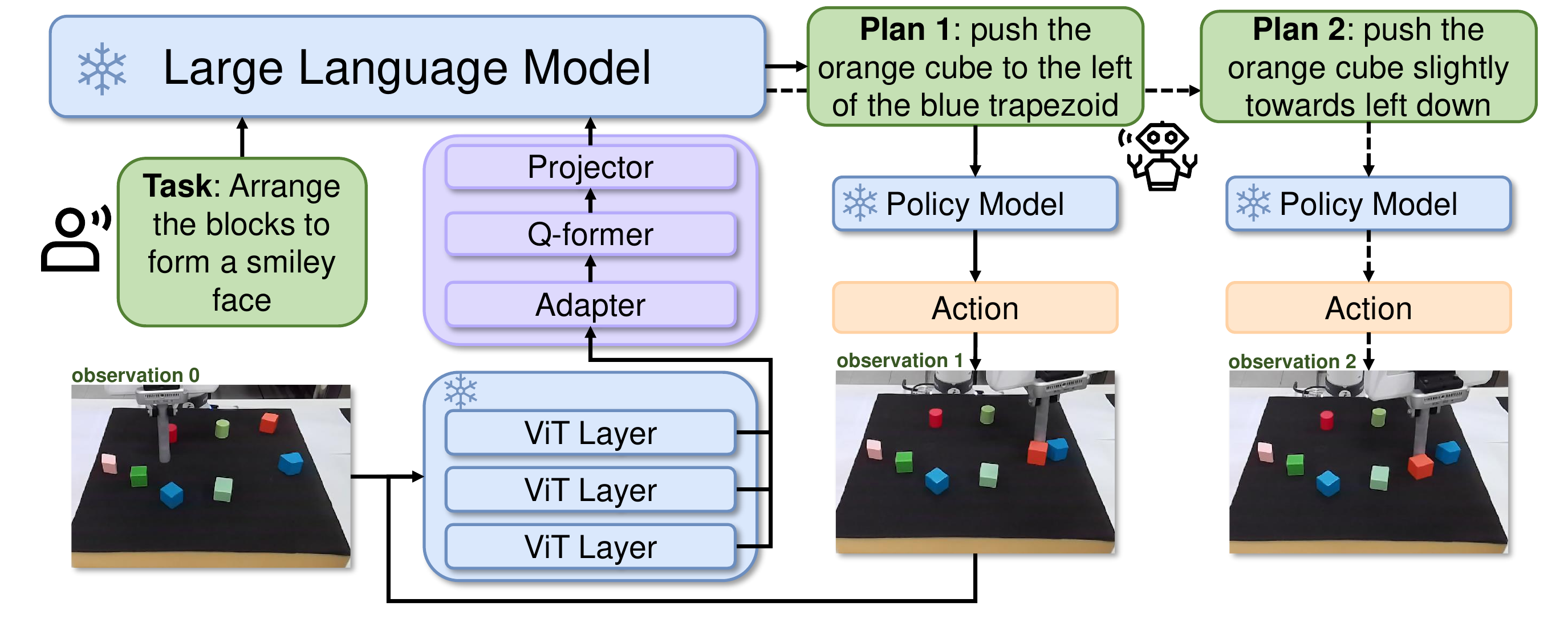}
\caption{
\textbf{CogLoop} consists of three main components. 
1) A pre-trained ViT that serves as efficient feature extractors.
2) Parameter-efficient tuning module includes a Vision Adapter and a combined Q-former with a projector to align multi-stage visual features with language space.
3) A frozen LLM which processes the task description and visual observation to reason out sub-task plans.
The plans are subsequently applied to a frozen Policy Model to generate actionable steps, which are then applied in embodied environments to obtain the next observation state. The dashed line indicates the next plan after generating the previous one. [Best viewed in color]}
\label{fig:model_architecture}
\end{figure}

\subsection{CogLoop as Robot Planner}\label{subsec:model_architecture}

In this section, we introduce \textbf{CogLoop}, an approach that integrates the generation of robotic sub-task plans using advanced pre-trained multi-modal large language models. To achieve this, we adopt a decoder architecture commonly used in multi-modal LLMs \cite{blip,minigpt4}. As shown in Figure~\ref{fig:model_architecture}, the vision modality is first encoded and then aligned to the language decoder space. The decoder then generates language-based robotic sub-task plans.

Drawing inspiration from the latest MiniGPT-4 model \cite{minigpt4}, which demonstrates exceptional vision-language understanding and reasoning capabilities, we adopt its unimodal model configuration and projector for robotic planning learning. Specifically, we utilize Vicuna \cite{vicuna} as our language decoder, trained by fine-tuning LLaMA \cite{llama} on user-shared conversations, and ViT-G/14 from EVA-CLIP \cite{evaclip} as our vision encoder. To synchronize visual features with pre-trained language encoders, we employ pre-trained Q-Former in BLIP-2 \cite{blip} and the projector in MiniGPT-4 \cite{minigpt4} as our visual embedding components.

The model parameters of MiniGPT-4 \cite{minigpt4} are openly available, allowing us to use them as the initialization parameters for our proposed \textbf{CogLoop}. To optimize these models, we focus on two primary objectives. First, we aim to preserve the inherent abilities of multi-modal LLMs, including visual perception, language generation, and few-shot transfer capabilities, in order to address potential challenges such as catastrophic forgetting \cite{kirkpatrick2017overcoming}. Additionally, we strive to generate reliable language-based sub-task plans through our further parameter-efficient fine-tuning. To achieve these objectives, we implement two key strategies. First, we maintain the frozen state of the unimodal pre-trained decoder and vision encoder throughout the training process. Second, we capitalize on the use of Vision Adapter to obtain more fine-grained visual features and Vision Tokenizer \cite{blip,minigpt4} to align visual features within the language decoder.

\textbf{Vision Adapter} \ Our scenario requires both local and global spatial perception (e.g., ``Is the blue moon in the upper left or lower left of the green cube nearby?'' and ``The specific position of the blue cube on the board'') as well as abstract semantic understanding (e.g., ``What does a smiley face look like and does it look like that now?''). Therefore, we introduce an additional Vision Adapter that integrates multi-stage features through attention mechanism and bottleneck design as follow:
\begin{equation}
    V_{ext} = \mathrm{Block}(V_{i}, V_{j}, V_{k}) + V_{k},
\label{eq:adapter}
\end{equation}
where $V_{ext}$ denotes the extracted features from Vision Adapter, $V_{i}$, $V_{j}$ and $V_{k}$ denote the output features of the ViT's $i$-th, $j$-th and $k$-th layer. The $\mathrm{Block}(\cdot, \cdot, \cdot)$ operation consists of multi-head attention operation defined in \cite{transformer} and a linear projection.
In this paper, we set $i$, $j$ and $k$ to 13, 26 and 39 for a 39-layered ViT.

\textbf{Vision Tokenizer} \ We fine-tune the Q-former in BLIP-2 \cite{blip} and the projector in MiniGPT-4 \cite{minigpt4} to capture visual features that are highly relevant to language-based task instructions. Specifically, the Vision Tokenizer can be formally defined as follows:
\begin{equation}
V_{emb} = \mathrm{Proj}(\mathrm{Q\text{-}former}\ (V_{ext})),
\end{equation}
where $V_{emb}$ represents the aligned visual embeddings that can be integrated into pre-trained LLMs.

\textbf{Objective} \ CogLoop generates sub-task plans in an auto-regressive manner. Our loss function can be formulated as follow:
\begin{equation} 
\min_{\theta} {\textstyle \sum_{  \boldsymbol{\mathcal{D}}}} \ \mathrm{CE}(\mathbf{p},\pi_{\theta} (l, o)),
\end{equation}
where $\mathbf{p}$ and $\mathrm{CE}(\cdot, \cdot)$ denote the generated plan and CrossEntropy loss function, respectively.
\section{Experiment}

In this section, we first show the training details in the experiment, then compare our model with four baseline methods, and finally analyze the experimental results from different perspectives.

\subsection{Training Details}

CogLoop is optimized via Adam \cite{adam} with a decoupled weight decay \cite{weight} of 0.05. The peak learning rate is 2e-5 and decays according to a cosine learning rate schedule. We use images of size 224x224 without any data augmentation. We train all models 2000 iterations with a batchsize of 8 in total. Our code is implemented on the PyTorch \cite{torch} toolbox on two NVIDIA RTX A6000 GPUs. 

\subsection{Experimental Settings}

\begin{figure}[t]
    \centering
    \includegraphics[width=0.95\linewidth]{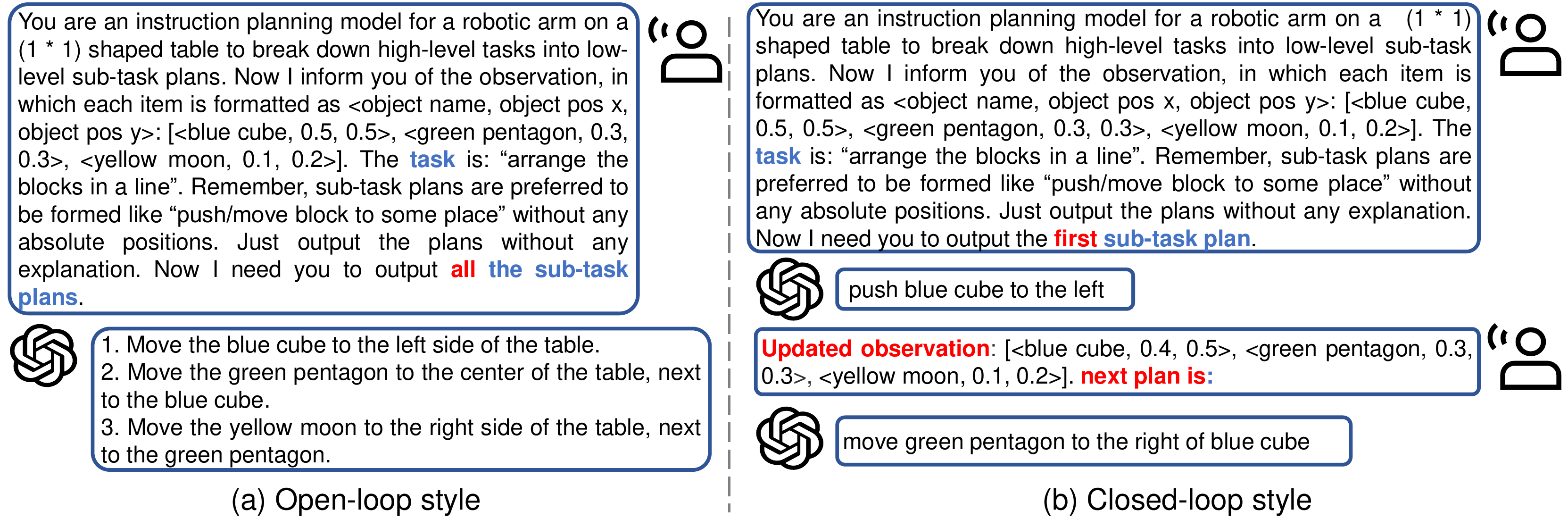}
    \caption{Examples of prompts and response from LLM in open-loop and closed-loop settings. We highlight importance in \textcolor{NavyBlue}{blue} and difference in \textcolor{red}{red}. [Better viewed in color.]}
    \label{fig:openclose}
\end{figure}

\paragraph{Compared Methods} 
We compare our model, CogLoop, with two mainstream pipelines, open-loop and closed-loop with language observation, that are commonly used in existing works for high-level cognitive plan. For LLM, we adopt two powerful LLMs (i.e., ChatGPT and GPT-4) for planning. In both settings, we first design a prompt~(shown in Appendix~\ref{app:baseline}) for LLM to include the description of the observed image with language and target output. A separate control model (e.g., LAVA \cite{lynch2022interactive}) is then leveraged to perform these sub-tasks. In open-loop setting, plans are generated in advance and are not updated without any feedback in this pipeline. While in closed-loop with language observation setting, prompt with instant feedback of observation are fed to update the planning. Examples of prompts for both settings are demonstrated in Figure \ref{fig:openclose}. To better convey spatial information to LLMs, we incorporate exact positions into the prompts to enable LLMs to better understand the context of the task at hand. 
Compared to these baseline methods, our model, \textbf{CogLoop}, employs a closed-loop approach and infuse visual perception into LLM for planning generation and updating. 

\paragraph{Evaluation and Metrics} 
For all tasks in the collected data, we divide them into an 80\% training set and a 20\% testing set. We evaluate the model by marking a high-level task as successful if the final state satisfied the high-level language instruction after a specific number of steps or plans. In particular, we compare the final position of the building block to its position in the ground truth data. If the distance between the two is less than a pre-defined threshold value (0.08 in our setting), the task is deemed successfully performed. We report the success rate of all models with the same number of sub-tasks (i.e., 15) and action steps in each sub-task (i.e., 10).

\begin{table}[h]\centering
\caption{Results of CogLoop as plan model compared to mainstream methods using LLM in simulator. All methods adopts LAVA \cite{lynch2022interactive} as the low-level action execution model}
\resizebox{\textwidth}{!}{
\begin{tabular}{lllc}
\hline
Example Model Type &  Open/Closed Loop & Re-Implemented Plan Model & Success Rate (\%)       \\  \hline
\{SayCan \cite{saycan},   & \multirow{2}{*}{open}  & ChatGPT        &        8.1                  \\ 
Grounded Decoding \cite{Grounded}\} &  & GPT-4        &    9.0  \\ \hline
\{Text2Motion \cite{text2motion},  & \multirow{2}{*}{closed w/ language}  &ChatGPT          &   2.1                    \\
 ChatGPT for Robotics \cite{chatgpt}\} &  & GPT-4          &  16.4  \\ \hline
Ours (CogLoop) & closed w/ vision  &Embodied robot model & \textbf{23.5}  \\ \hline
\end{tabular}}
\label{tab:main_exp}
\end{table}

\subsection{Results and Analysis}
Table \ref{tab:main_exp} presents the results of our model in comparison with other baseline methods in a simulated environment. We analyze the results to address the following three questions:
\begin{itemize}
\setlength{\itemsep}{0pt}
\setlength{\parsep}{0pt}
\setlength{\parskip}{0pt}
\item Does the collected AlphaBlock dataset prove to be useful?
\item In terms of high-level planning tasks, which approach is superior: open-loop or closed-loop?
\item Which modality offers better feedback for plan updates: language or image?
\end{itemize}

\paragraph{AlphaBlock is valuable for learning high-level robotic perception and reasoning capabilities.} As demonstrated in the results (Table \ref{tab:main_exp}), our model trained on the AlphaBlock dataset outperforms those that directly output plans using LLMs, showcasing the dataset's potential in developing more sophisticated robotic systems.

\paragraph{Closed-loop with plan updating proves superior.}
As evidenced by the results in Table \ref{tab:main_exp}, closed-loop settings, which take into account the current observation after executing previous actions when generating sub-task plans, significantly outperform open-loop settings except for ChatGPT planning. This observation aligns with findings from prior research studies \cite{palme,text2motion,chatgpt}. The underlying reason is intuitive, as interacting with the physical world may result in various states. A closed-loop setting is adept at detecting unanticipated changes, enabling real-time updates to sub-task plans. The poor performance of ChatGPT in a closed-loop setting with language observation can be attributed to its lack of spatial perception and reasoning capabilities. Since ChatGPT is primarily designed for natural language processing tasks, it may struggle in scenarios that demand a deeper understanding of spatial relationships and reasoning, which are essential for many robotic tasks.

\paragraph{Vision enhances perception for plan updating.}
In a closed-loop setting, our CogLoop, which employs visual perception for plan updating, demonstrates a significant improvement when compared to other planners that utilize LLM with language observation. Specifically, our approach outperforms ChatGPT and GPT-4 with significant improvements of 21.4\% and 7.1\%, respectively, even when the exact positions of each block are provided to the LLMs. This result suggests that even with a less powerful generative model (i.e., miniGPT-4), our model effectively integrates visual signals into the LLM, resulting in superior sub-task planning. Moreover, this highlights that real-time observation at each step in CogLoop offers valuable feedback for immediate plan adjustments in subsequent steps.

\subsection{Ablation Studies}
We carried out ablation studies to investigate the impact of total steps and steps for re-planning. Additionally, we examined the effectiveness of various components within our model design. 

\begin{table}[h]\centering
\caption{Ablation study on step length of each high-level task during inference time}
\begin{tabular}{ccc}
\hline
Total Steps  &  Steps/Re-plan & Success Rate \\ \hline
50 &      10       &    8.5                      \\
100 & 10 & 18.6 \\
150 & 10 & 23.5  \\
200 &     10   &     23.7                          \\ 
\hline
\end{tabular}
\label{tab:ab_step}
\end{table}
\paragraph{Does an increased number of steps lead to a higher success rate?}
To examine the optimal number of steps for high-level tasks, we conduct experiments with varying step counts during inference time: 50, 100, 150, and 200, respectively. The results shown in Table \ref{tab:ab_step} reveal that with an increasing number of steps, the success rate gradually rises and converge at step 150. This indicates that a higher number of steps can enhance the success rate, though it converges at a certain point.

\begin{table}[h]\centering
\caption{Ablation study on re-planning frequency for each high-level task during inference time}
\begin{tabular}{cccc}
\hline
Total Steps  &  Steps/Re-plan & Success Rate & Average Time/Task (s) \\ \hline
150  &     5        &  24.1  &    82.9                  \\
150 & 10 &  23.5 & 49.3 \\
150  &     15      &  21.2  &   38.2                  \\
\hline
\end{tabular}
\label{tab:ab_frequency}
\end{table}

\paragraph{What is the optimal number of steps for updating plans in high-level tasks?}
The results shown in Table \ref{tab:ab_frequency} indicate that updating the plan more frequently leads to a higher success rate. However, more frequent updates also result in a substantial increase in computational burden. To balance effectiveness and efficiency, we opted to perform re-planning every 10 steps. This approach allows for better adaptation to changes in the environment and task requirements without significantly impacting computational resources.

\begin{table}[h]\centering
\caption{Ablation study on model design of frozen and tuned part. Note that we keep Vision Projector tuned for instruction tuning to match the policy input pattern}
\begin{tabular}{ccc}
\hline
 Vision Q-former &  Adapter &  Success Rate       \\ \hline
\begin{minipage}[b]{0.02\columnwidth}
        \centering
        \raisebox{-.5\height}{\includegraphics[width=\linewidth]{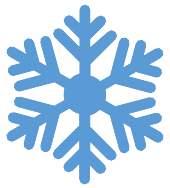}}
    \end{minipage} 
    &                 &    20.8      \\
\begin{minipage}[b]{0.02\columnwidth}
        \centering
        \raisebox{-.5\height}{\includegraphics[width=\linewidth]{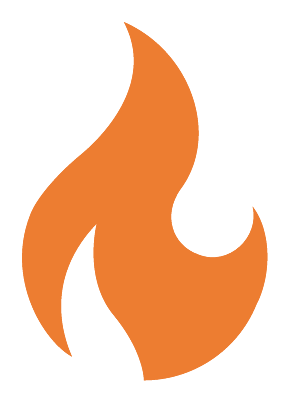}}
    \end{minipage} 
    &                   &   21.4    \\ 
\begin{minipage}[b]{0.02\columnwidth}
        \centering
        \raisebox{-.5\height}{\includegraphics[width=\linewidth]{figs/fire.pdf}}
    \end{minipage} 
    &     \checkmark              &    23.5    \\\hline
\end{tabular}
\label{tab:ab_component}
\end{table}

\paragraph{What is the effectiveness of tuning Vision Q-former and incorporating the Adapter component?}
To assess the impact of freezing or tuning Vision Q-former and the effectiveness of our proposed Adapter in the model design, we devise various settings as presented in Table \ref{tab:ab_component}. The results indicate a substantial improvement in performance when Vision Q-former is made learnable. This is expected, as the original Q-former is trained to fit the language model OPT \cite{zhang2022opt}, while we aim to adapt it to Vicuna \cite{vicuna}. This demonstrates the necessity of tuning the modality alignment when the target language model differs. 
Moreover, our specifically designed Adapter plays a significant role in improving performance, indicating that the standard vision backbone might not be directly suitable for robotic tasks. However, our Adapter effectively enhances the spatial reasoning capability necessary for robotic tasks. The best performance is achieved when both design elements, the Vision Q-former and the Adapter, are utilized in tandem. This highlights the importance of tailoring model components to cater to the unique requirements of robotic applications.

\subsection{Real-World Deployment}
In order to further validate the effectiveness of our model, we deploy it on a robot arm within a real-world environment after training it in the simulator. During real-world testing, we utilize a Franka Emika Research 3 robot arm. We establish a table-top setup and employ a Kinect DK camera to capture RGB images of the scene, closely resembling the simulator environment. To bridge the gap between physical and virtual building blocks, we make use of an image segmentation model \cite{kirillov2023segment} to substitute the physical blocks with their digital counterparts from the simulator. We select one task from each task family to demonstrate the effectiveness of our model in a real-world environment. The demonstrations are shown in \small{\url{https://www.youtube.com/watch?v=ayAzID1_qQk}}.
\section{Conclusion}
In this paper, we have presented a novel robot learning framework that addresses the challenging tasks associated with the disconnect between high-level instructions and low-level actions in robot manipulation tasks. By collecting one of the first cognitive robot building block dataset, our framework empowers robots to generate plans and execute them autonomously in the physical world, such as creating a smiley face, or making a letter ``Q'' using building blocks. To validate the effectiveness of our framework, we conduct extensive experiments on 35 challenging tabletop manipulation tasks in both simulated and real-world environments, and observe clear performance increases. By seamlessly integrating high-level cognitive capabilities into robot manipulation tasks, our framework opens up new possibilities for more efficient and intelligent interaction with the physical world. This advancement holds great potential for a wide range of applications, where robots can understand and execute high-level instructions more easily. As for future work, we plan to further refine our framework by training more capable low-level action execution model, and explore its application in more complex and diverse scenarios (e.g., grasping 3D objects).

%%%%%%%%%%%%%%%%%%%%%%%%%%%%%%%%%%%%%%%%%%%%%%%%%%%%%%%%%%%%
{\small
\bibliographystyle{plain}
\bibliography{arxiv}
}
%%%%%%%%%%%%%%%%%%%%%%%%%%%%%%%%%%%%%%%%%%%%%%%%%%%%%%%%%%%%

\appendix
\section{Prompt for Layout Collection} \label{app:layout}

\subsection{Details of our prompt}
\begin{figure}
    \centering
    \includegraphics[width=1.0\linewidth]{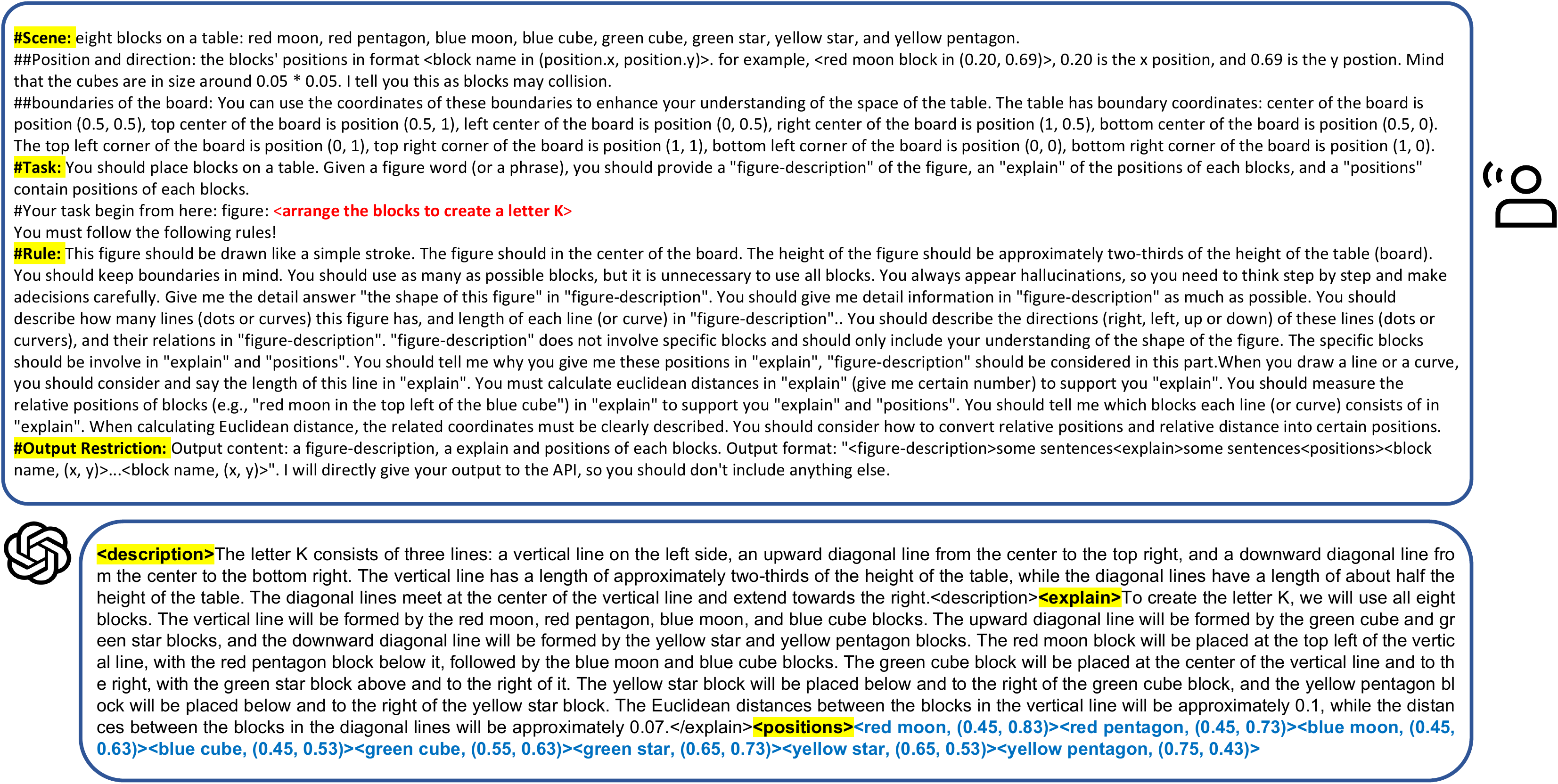}
    \caption{The prompt for GPT-4 to generate the position of the given layout letter ``K''.}
    \label{fig:layout_prompt}
\end{figure}

We show one prompt example for layout collection in Figure~\ref{fig:layout_prompt}. The hyper-parameters of GPT-4 are shown in Table~\ref{tab:ab_hyper}.

\begin{table}[h]\centering
\caption{The hyper-parameters of GPT-4.}
\begin{tabular}{ccc}
\hline
 Parameter  & Value \\ \hline
Temperature &  0.22 \\
Top-P & 0.95 \\
\hline
\end{tabular}
\label{tab:ab_hyper}
\end{table}

\subsection{Effectiveness of <description> and <explain>}

To verify the effectiveness of  $description$ and $explain$, we remove the sections related to $description$ and $explain$ in our prompt and only ask GPT-4 to provide coordinates. We manually filter 50 groups of coordinates, and the retention rate of the layout is shown in Table~\ref{tab:ab_cot}.
\begin{table}[h]\centering
\caption{The comparison of the retaining rate of whether use description and explain in the prompt.}
\begin{tabular}{ccc}
\hline
 Description  & Explain & Retain rate\\ \hline
 &   & 12\% \\
\checkmark & \checkmark & 25\%\\
\hline
\end{tabular}
\label{tab:ab_cot}
\end{table}

\begin{figure}
    \centering
    \includegraphics[width=0.9\linewidth]{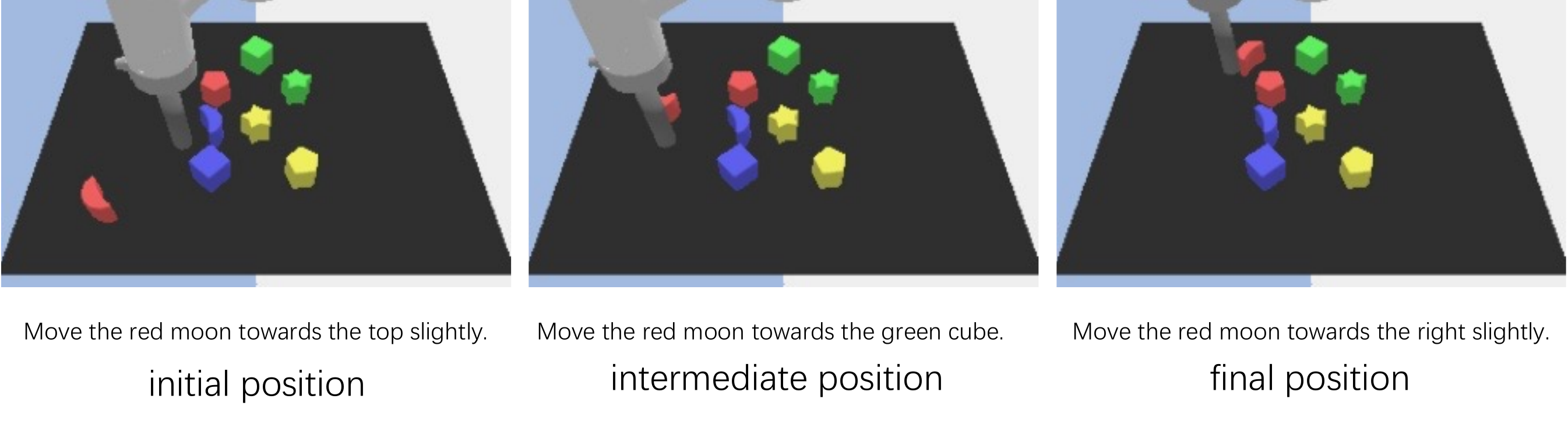}
    \caption{The task of moving the red moon from position (0.17, 0.40) to position (0.76, 0.17).}
    \label{fig:collect_planning}
\end{figure}

\subsection{Effectiveness of self-verification approach}
To verify the effectiveness of our self-verification approach, we manually rank 50 pairs of our data.
As shown in Table~\ref{tab:verify_rank}, 16\% of the second response has been improved, and 86\% of the second response is no worse than the first one. This demonstrates the effectiveness of our self-verification method.

\begin{table}[h]\centering
\caption{Manually annotated rankings of 50 self-verification pairs, ``first'' means the origin response 1, and ``second'' means the response 2 after ask GPT-4 ``Are you sure your answer is correct?''.}
\begin{tabular}{ccc}
\hline
 Rank & Quantity & Ratio \\ \hline
Second > First & 16 & 32\% \\
Second = First & 27 & 54\% \\
Second < First & 7 & 14\% \\
\hline
\end{tabular}
\label{tab:verify_rank}
\end{table}

\section{Prompt for Real-Time Planning Collection}\label{app:coll_real}
\subsection{Details of our prompt}
\begin{figure}
    \centering
    \includegraphics[width=1.0\linewidth]{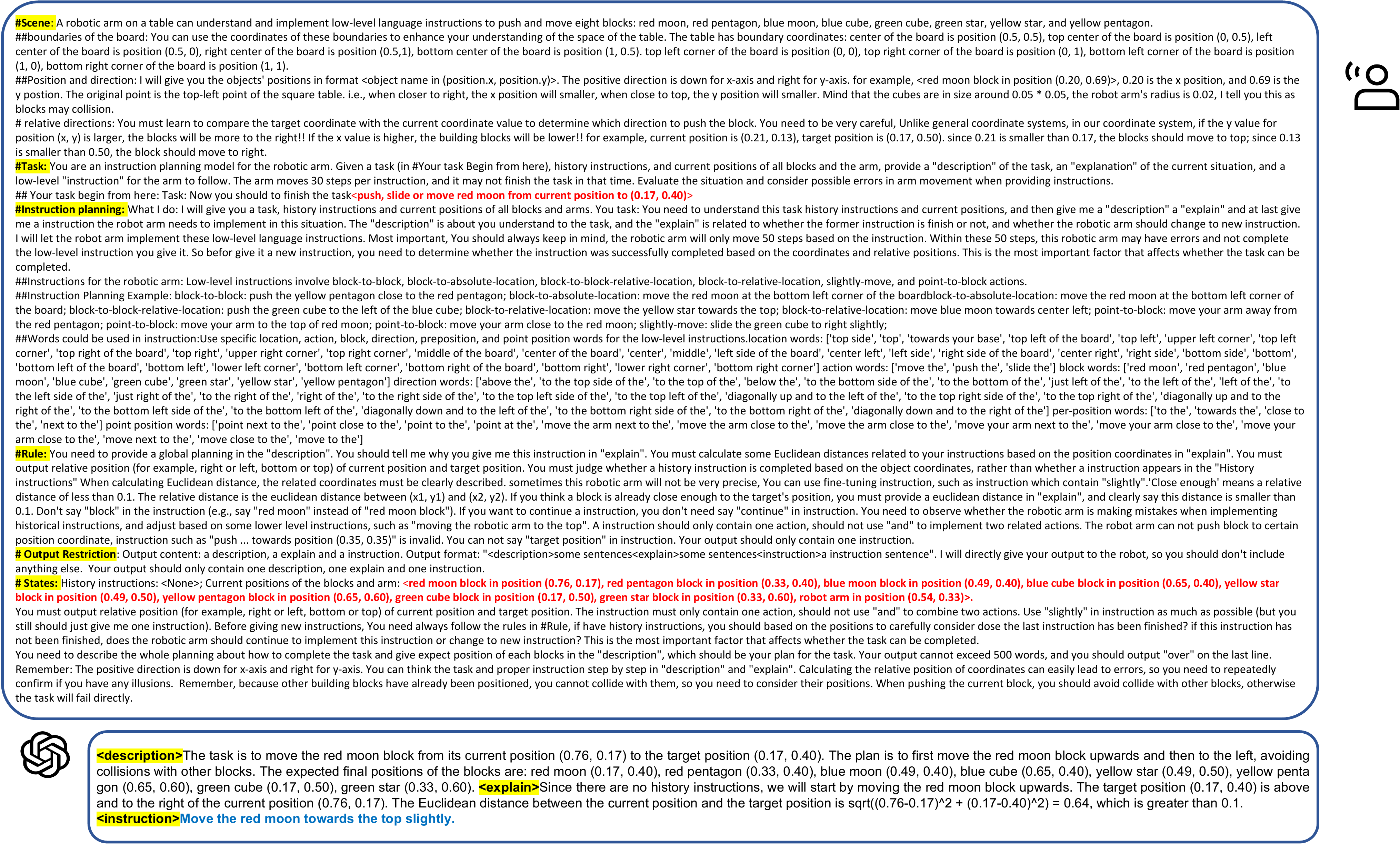}
    \caption{The prompt for GPT-4 to move the ``red moon'' to target position (0.17, 0.40).}
    \label{fig:pos_prompt}
\end{figure}

This prompt is used to let the GPT-4 model provide a real-time plan to move the certain block to certain positions. such as ``move the red moon to position (0.76, 0.17)''. We provide our prompt, as shown in Figure~\ref{fig:pos_prompt}.

\subsection{Examples of real-time planning}
We provide real-time planning example for the target position movement task, as shown in Figure~\ref{fig:collect_planning}.

\section{Prompt for Baseline Methods in Real-Time} \label{app:baseline}
\subsection{Details of our prompt}
This prompt is used for our high-level task real-time planning baseline, such as ``move the red moon to form the letter R.'' Different from the prompt used to collect the real-time planning~(mentioned in section~\ref{app:coll_real}), we are not providing the target position of the red moon. This setting requires the model to have the cognitive ability to infer the possible target position to form the letter `R'. This prompt is similar to the real-time planning collection prompt~(shown in Figure~\ref{fig:pos_prompt}), the difference is the task and some rules.

\subsection{Example of <description> and <explain>}
\begin{figure}
    \centering
    \includegraphics[width=0.9\linewidth]{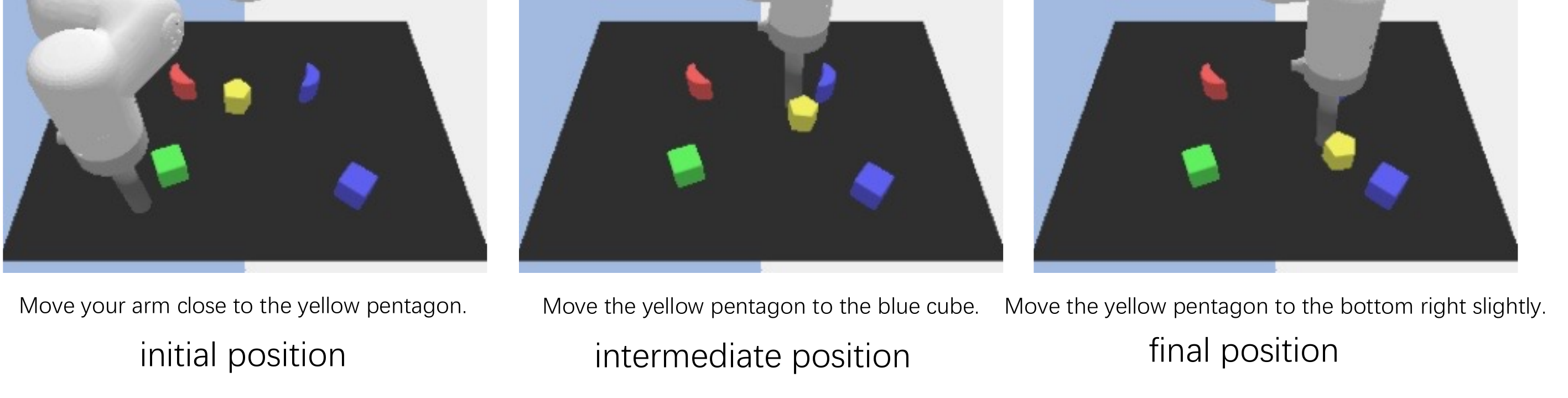}
    \caption{The high level task of move the ``yellow pentagon'' to form the letter `Q'.}
    \label{fig:baseline_realtime}
\end{figure}

We provide an example response from GPT-4 about how to move the ``yellow pentagon'' to form the letter `Q', the corresponding images are shown in Figure~\ref{fig:baseline_realtime}.
\begin{enumerate}
    \item To create a letter Q using the yellow pentagon, we need to move the yellow pentagon to the bottom right corner of the board, and then slightly move it up and to the left. The expected final position of the yellow pentagon is around (0.9, 0.9).
    \item Since there are no history instructions, we will start by moving the robotic arm close to the yellow pentagon. The current position of the yellow pentagon is (0.39, 0.48), and the robotic arm is at (0.79, 0.27).
    \item Move your arm close to the yellow pentagon.
\end{enumerate}

% \label{subsec:data_collection}
\begin{figure}[htbp]
    \centering
    \includegraphics[width=\linewidth]{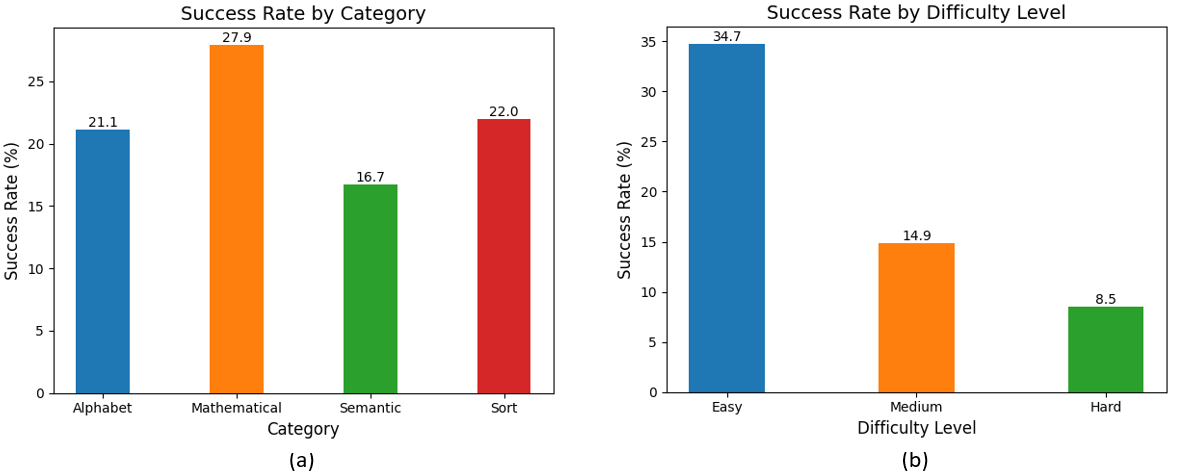}
    \caption{Task success rate statistics in the simulator environment.}
    \label{fig:fenxi}
\end{figure}

\section{Task Success Rate Statistics in simulator environment}

We further report the average sample success rate of all tasks based on the category of task family and level of difficulty.

\subsection{Results of different task families}
We report the success rate of each task family categorized in our main paper in Figure \ref{fig:fenxi}(a). Experimental results show that Mathematical Geometries generally achieve better results, because their layout is usually relatively simpler, and the control model is easier to execute successfully.

\subsection{Results on different difficulty}

We divide all the tasks into three difficulty levels based on layout complexity and ambiguity. The specific divisions are as follows.
\begin{itemize}
    \item \textbf{Easy}: horizontal lines, vertical lines, triangles, squares, `C', `D', `F', `H', `I', `L', `T', `U', `V'.
    \item \textbf{Medium}: sort by color, sort by shape, circles, `A', `E', `J', `O', `Q', `R', `W', `X'.
    \item \textbf{Hard}: smiley faces, smiley faces with green eyes, `B', `G', `K', `M', `N', `P', `S', `Y', `Z'.
\end{itemize}

We report the success rate based on difficulty level in Figure \ref{fig:fenxi}(b). Experimental results show that layout complexity and ambiguity indeed pose significant challenges to cognitive planning models.

\end{document}